\documentclass[conference,a4paper]{IEEEtran}
\IEEEoverridecommandlockouts
\usepackage{cite}
\usepackage{amsmath,amssymb,amsfonts}
\usepackage{algorithmic}
\usepackage{graphicx}
\usepackage{textcomp}
\usepackage{xcolor}
\usepackage[colorlinks=true, urlcolor=blue, citecolor=blue, linkcolor=darkgray]{hyperref}
\renewcommand\figureautorefname{Fig.}
\renewcommand\tableautorefname{Tab.}
\def\BibTeX{{\rm B\kern-.05em{\sc i\kern-.025em b}\kern-.08em
    T\kern-.1667em\lower.7ex\hbox{E}\kern-.125emX}}

\usepackage{amsmath,amssymb,amsfonts}
\usepackage{textcomp}
\usepackage{xcolor}
\usepackage{float}
\usepackage{url}
\usepackage{booktabs}
\usepackage{caption}
\usepackage{stfloats}
\usepackage{makecell}
\graphicspath{{figures/}}
\labelformat{equation}{\textup{(#1)}}
\usepackage{flushend}
\usepackage{tikz}
\usepackage{multirow}
\usepackage{todonotes}

\usepackage{array}
\newcolumntype{L}[1]{>{\raggedright\let\newline\\\arraybackslash\hspace{0pt}}m{#1}}
\newcolumntype{C}[1]{>{\centering\let\newline\\\arraybackslash\hspace{0pt}}m{#1}}
\newcolumntype{R}[1]{>{\raggedleft\let\newline\\\arraybackslash\hspace{0pt}}m{#1}}

\def\usenatbib{1}
\ifx\usenatbib\undefined
    \usepackage{cite}
\else
    \makeatletter
    \let\NAT@parse\undefined
    \makeatother
    \usepackage[numbers,sort]{natbib}
    \def\refname{References}

    \setcitestyle{citesep={], [}}
    \makeatletter
    \def\NAT@def@citea{\def\@citea{\NAT@separator}}%
    \makeatother
\fi

\graphicspath{ {./images/} }

\let\orgautoref\autoref

\renewcommand{\autoref}
        {\def\equationautorefname{Equation}%
         \def\figureautorefname{Fig.}%
         \def\subfigureautorefname{Fig.}%
         \def\Itemautorefname{item}%
         \def\tableautorefname{Table}%
         \def\exerciseautorefname{Exercise}%
         \def\starexerciseautorefname{Exercise}%
         \def\sectionautorefname{Section}%
         \def\subsectionautorefname{Section}%
         \def\subsubsectionautorefname{Section}%
         \def\chapterautorefname{Section}%
         \def\partautorefname{Part}%
         \orgautoref}

\usepackage{hyperref}
\hypersetup{
colorlinks=true,
urlcolor=red,
}
\begin{document}

\title{Are Time-Series Foundation Models Ready for E-Nose Data? An Empirical Assessment of Their Embeddings \\
}

\author{
\IEEEauthorblockN{Taeyeong Choi}
\IEEEauthorblockA{\textit{Department of Information Technology} \\
\textit{Kennesaw State University}\\
Marietta, GA 30060, USA \\
tchoi3@kennesaw.edu}
\and
\IEEEauthorblockN{Mohammed Kamruzzaman}
\IEEEauthorblockA{\textit{Department of Agricultural and Biological Engineering} \\
\textit{University of Illinois Urbana-Champaign}\\
Urbana, IL 61801, USA \\
mkamruz1@illinois.edu}
}

\maketitle

\begin{abstract}
Inspired by advances in natural language processing and computer vision, ``time-series foundation models''~(TSFMs) have recently been introduced with the promise of strong generalization across diverse time-series tasks, including forecasting, classification, and anomaly detection, as well as across domains such as healthcare, climate science, and manufacturing. 
However, their utility for \emph{gas-sensing} data remains largely unexplored. 
To address this gap, this paper systematically evaluates recent TSFMs on electronic nose (E-Nose) data. 
In particular, we investigate whether embeddings produced by representative TSFMs, including \mbox{Chronos-2} and MOMENT, provide effective representations for gas identification and concentration prediction. 
Specifically, we show that fine-tuning is necessary to achieve satisfactory performance on E-Nose data, and fusing TSFM embeddings with representations learned by specialized predictive models can further improve the performance, suggesting both the potential and limitations of current TSFMs for gas-sensing applications.

\end{abstract}


\section{Introduction}

Advanced time-series foundation models (TSFMs) have recently been introduced to develop general-purpose models capable of supporting predictive tasks across diverse time-series domains~\citep{moment, Chronos-2, GCM23}. 
By pretraining on large-scale collections of time-series data, these models are expected to learn reusable temporal representations that can be applied to downstream tasks in zero-shot settings or adapted through fine-tuning with limited task-specific data. 
Recent studies have reported promising results in domains such as healthcare~\citep{ZSZVH25, LDXFSHSSYY26}, climate systems~\citep{RSSPW25}, and manufacturing~\citep{ZFJS25}, suggesting that TSFMs, including Chronos-2~\citep{Chronos-2} and MOMENT~\citep{moment}, can capture transferable temporal patterns from pretraining data and generalize to previously unseen time-series data.

However, the generalizability of TSFMs to Electronic Nose (E-Nose) data remains largely unexplored, while E-Nose systems have been widely adopted to detect gases and odors and estimate concentration levels in practical domains such as environment and food safety monitoring~\citep{RB14, SZFGS14}. 
In E-Nose, arrays of cross-reactive metal-oxide (MOX) sensors are configured to exhibit different sensitivity patterns toward target gases, including volatile organic compounds (VOCs), thereby producing cross-sensor response patterns for gas analysis. 
As a result, the generated data are inherently multivariate and temporal, making them a natural candidate for TSFMs. 

\begin{figure}[t]
    \centering
    \includegraphics[width=.90\linewidth]{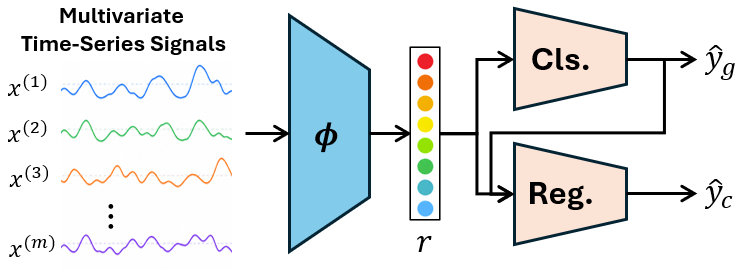}
    \caption{
    Illustration of the designed pipeline used to examine the effectiveness of embeddings~$r$ extracted by each time-series encoder~$\phi$ (e.g.,~\mbox{Chronos-2}).
    Classification and regression heads are trained on~$r$ to predict the gas type~$y_g$ and concentration~$y_c$, respectively.} 
    \label{fig:concept}
\end{figure}

To address this gap, 
we investigate whether embeddings produced by pretrained TSFMs encode informative representations of E-Nose time-series signals (cf.~\autoref{fig:concept}). 
We select Chronos-2~\citep{Chronos-2} and MOMENT~\citep{moment} as representative TSFMs and use them as encoders, followed by task-specific prediction heads. 
The evaluation is conducted in a practical multi-task scenario involving gas-type identification as a classification task and concentration-level prediction as a regression task.

To the best of our knowledge, this is the first study to investigate TSFMs for E-Nose sensor time-series data.
Since publicly available descriptions of these foundation models do \emph{not} indicate pretraining on chemical gas sensor data, this study examines their out-of-domain generalizability and transferability to gas-sensing tasks.
We design experiments to address the following research questions:
\begin{enumerate}
\item Do pretrained TSFMs offer useful embeddings for gas-type and concentration prediction without fine-tuning?
\item Can TSFM-based embeddings outperform domain-specific models for gas sensing?
\item Does combining TSFM-based embeddings with domain-specific ones further improve performance?
\end{enumerate}


\begin{table*}[t]
    \small
    \centering
    \begin{tabular}{|C{15mm}|C{15mm}||C{11mm}|C{11mm}|C{11mm}|C{11mm}|C{11mm}|C{11mm}|C{11mm}|C{11mm}|C{11mm}|C{11mm}|}
    \hline
     \multicolumn{2}{|c||}{} & \multicolumn{2}{c|}{Weighted Avg.} \normalsize &  \multicolumn{2}{c|}{B1} \normalsize & \multicolumn{2}{c|}{B2} \normalsize & \multicolumn{2}{c|}{B3} \normalsize & \multicolumn{2}{c|}{B4} \normalsize \\
        \cline{3-12}
     \multicolumn{2}{|c||}{} & Acc. & F1 & Acc. & F1 & Acc & F1 & Acc. & F1 & Acc. & F1  \\
    \hline
    \hline
    \multicolumn{2}{|c||}{MLP} & $.704$  & $.654$ & $\mathbf{.896}$\newline\tiny$\pm.028$ & $\mathbf{.894}$\newline\tiny$\pm.028$ & $\mathbf{.923}$\newline\tiny$\pm.034$ & $\mathbf{.918}$\newline\tiny$\pm.037$ & $.517$\newline\tiny$\pm.028$ & $.393$\newline\tiny$\pm.037$ & $.258$\newline\tiny$\pm.006$ & $.167$\newline\tiny$\pm.015$  \\
    \hline    \multicolumn{2}{|c||}{DBFE}  & $.667$  & $.632$& $.507$\newline\tiny$\pm.033$ & $.472$\newline\tiny$\pm.056$ & $.777$\newline\tiny$\pm.055$ & $.747$\newline\tiny$\pm.081$ & $.735$\newline\tiny$\pm.124$ & $.714$\newline\tiny$\pm.148$ & $.629$\newline\tiny$\pm.142$ & $.556$\newline\tiny$\pm.143$ \\
    \hline
    \multirow{2}{*}{Chronos-2} & Frozen & $.496$ & $.404$ & $.479$\newline\tiny$\pm.015$ & $.321$\newline\tiny$\pm.009$ & $.579$\newline\tiny$\pm.093$ & $.552$\newline\tiny$\pm.115$ & $.271$\newline\tiny$\pm.018$ & $.139$\newline\tiny$\pm.031$ & $.817$\newline\tiny$\pm.075$ & $.802$\newline\tiny$\pm.100$\\
    \cline{2-12}
     & Fine-Tuned & $.845$  & $.839$ & $.881$\newline\tiny$\pm.036$ & $.879$\newline\tiny$\pm.038$ & $.813$\newline\tiny$\pm.015$ & $.806$\newline\tiny$\pm.014$ & $\mathbf{.890}
     $\newline\tiny$\pm.031$ & $\mathbf{.891}$\newline\tiny$\pm.029$ & $.750$\newline\tiny$\pm.142$ & $.723$\newline\tiny$\pm.170$ \\
     \hline
     \multirow{2}{*}{MOMENT} & Frozen & $.474$ & $.387$ & $.602$\newline\tiny$\pm.016$ & $.515 $\newline\tiny$\pm.016$ & $.583$\newline\tiny$\pm.046$ & $.515$\newline\tiny$\pm.042$ & $.296$\newline\tiny$\pm.018$ & $.191$\newline\tiny$\pm.027$ & $.354$\newline\tiny$\pm.006$ & $.271$\newline\tiny$\pm.009$  \\
     \cline{2-12}
     & Fine-Tuned & $.453$ & $.361$ & $.619$\newline\tiny$\pm.023$ & $.558$\newline\tiny$\pm.030$ & $.379$\newline\tiny$\pm.033$ & $.289$\newline\tiny$\pm.042$ & $.356$\newline\tiny$\pm.066$ & $.245$\newline\tiny$\pm.037$ & $.463$\newline\tiny$\pm.031$ & $.341$\newline\tiny$\pm.012$ \\
     \hline
     \multicolumn{2}{|c||}{MLP + Chronos-2} & $\mathbf{.876}$ & $\mathbf{.876}$ & $.888$\newline\tiny$\pm.005$ & $.887$\newline\tiny$\pm.004$ & $.854$\newline\tiny$\pm.029$ & $.845$\newline\tiny$\pm.032$ & $.888$\newline\tiny$\pm.084$ & $.884$\newline\tiny$\pm.090$ & $\mathbf{.875}$\newline\tiny$\pm.037$ & $\mathbf{.871}$\newline\tiny$\pm.004$ \\
    \hline
    \end{tabular}
    \caption{Gas-type identification performance for each held-out sensing unit (B1--B4). Overall results are reported as sample-weighted averages across sensing units, and smaller values below each unit-wise result denote standard deviations across three random splits.} 
    \label{tab:cls_results}
\end{table*}


\section{Study Design}

\subsection{Problem Formulation}

We formulate a practical multi-task learning problem in which a model simultaneously performs gas-type identification and concentration prediction. 
Formally, each sample is represented as a multivariate time-series matrix with $\tau$~time steps: $\mathbf{X} = [\mathbf{x}_{1}, \mathbf{x}_{2}, \ldots, \mathbf{x}_{\tau}]^\top \in \mathbb{R}^{\tau \times m}$, where $\mathbf{x}_{t} = [x^{(1)}_t, x^{(2)}_t, \ldots, x^{(m)}_t] \in \mathbb{R}^{m}$ denotes the multichannel response vector captured by the $m$~MOX sensors at time step $t$. 
As in~\autoref{fig:concept}, an encoder~$\phi$ maps each sample $\mathbf{X}$ to a compact vector embedding~$r \in \mathbb{R}^d$, which is then processed by a classifier~$\mathcal{F}$ for identification of the gas type~$y_g \in \{0, , ..., n-1\}$ and a regressor~$\mathcal{G}$ for prediction of the concentration~$y_c \in \mathbb{R}$, where $n$~is the number of available gas types.
The regressor additionally uses the output of $\mathcal{F}$ to condition its concentration prediction on the predicted gas type.

The performance of both classification and regression tasks is evaluated to determine whether each encoder captures informative patterns in the learned representation, which serves as the main input to the task-specific prediction heads.

\subsection{TSFM-Based Encoding}

In this work, we assess the capability of TSFMs as the encoder~$\phi$. 
Specifically, we employ \mbox{Chronos-2}~\citep{Chronos-2} and MOMENT~\citep{moment} to investigate whether their embedding spaces, learned from large and diverse time-series datasets, can produce informative representations~$r$ for gas-sensing tasks.

Chronos-2 was pretrained for forecasting on synthetic time-series datasets, using both temporal attention, which captures dependencies across time within each channel, and group attention, which enables information sharing across different channels.
In contrast, MOMENT was pretrained on Time-Series Pile~\citep{moment}, a large collection of public time-series datasets, for reconstruction of randomly masked patch sequences 
from the remaining context. 

Both Chronos-2 and MOMENT first divide the input sequence into non-overlapping temporal patches, each of which is finally mapped to a representation~$r$ through a series of network, including Transformer-based encoders, and pooling~\citep{Chronos-2, moment}. 
To obtain a compact representation for downstream tasks (i.e.,~gas-type and concentration prediction), we apply global average pooling across both the sensor (variable) and temporal patch dimensions, resulting in a fixed-dimensional representation $r \in \mathbb{R}^d$ for each input sample.


\begin{table*}[t]
    \small
    \centering
    \begin{tabular}{|C{22mm}|C{22mm}||C{11mm}|C{11mm}|C{11mm}|C{11mm}|C{11mm}|C{11mm}|C{11mm}|C{11mm}|C{11mm}|C{11mm}|}
    \hline
     \multicolumn{2}{|c||}{} & \multicolumn{2}{c|}{Weighted Avg.} \normalsize &  \multicolumn{2}{c|}{B1} \normalsize & \multicolumn{2}{c|}{B2} \normalsize & \multicolumn{2}{c|}{B3} \normalsize & \multicolumn{2}{c|}{B4} \normalsize \\
        \cline{3-12}
     \multicolumn{2}{|c||}{} & RMSE & MAPE & RMSE & MAPE & RMSE & MAPE & RMSE & MAPE & RMSE & MAPE \\
    \hline
    \hline
    \multicolumn{2}{|c||}{MLP} & $71.19$  & $76.60$ & $44.86$\newline\tiny$\pm4.30$ & $\mathbf{31.85}$\newline\tiny$\pm.53$ & $84.60$\newline\tiny$\pm.92$ & $\mathbf{63.98}$\newline\tiny$\pm.54$ & $80.26$\newline\tiny$\pm.8.67$ & $137.68$\newline\tiny$\pm19.09$ & $78.88$\newline\tiny$\pm4.48$ & $\mathbf{69.13}$\newline\tiny$\pm9.09$  \\
    \hline
    \multicolumn{2}{|c||}{Chronos-2 Fine-Tuned} & $61.95$  & $70.81$ & $61.08$\newline\tiny$\pm10.37$ & $61.65$\newline\tiny$\pm6.43$ & $74.90$\newline\tiny$\pm2.79$ & $76.65$\newline\tiny$\pm6.06$ & $48.67$\newline\tiny$\pm2.35$ & $59.21$\newline\tiny$\pm12.63$ & $\mathbf{64.35}$\newline\tiny$\pm8.92$ & $100.65$\newline\tiny$\pm6.20$  \\
     \hline
     \multicolumn{2}{|c||}{MLP + Chronos-2} & $\mathbf{48.52}$  & $\mathbf{57.21}$ & $\mathbf{43.57}$\newline\tiny$\pm2.01$ & $32.20$\newline\tiny$\pm4.00$ & $\mathbf{54.93}$\newline\tiny$\pm3.48$ & $73.11$\newline\tiny$\pm10.57$ & $\mathbf{35.83}$\newline\tiny$\pm6.38$ & $\mathbf{48.06}$\newline\tiny$\pm12.67$ & $70.95$\newline\tiny$\pm5.03$ & $93.72$\newline\tiny$\pm5.72$ \\
    \hline
    \end{tabular}
    \caption{Concentration prediction performance for each held-out sensing unit (B1--B4). The reporting format follows~\autoref{tab:cls_results}.}
    \label{tab:reg_results}
\end{table*}

\section{Experiments}

\subsection{Gas-Sensing Dataset}
\label{sec:dataset}

We use the Twin Gas Sensor Arrays dataset~\citep{FFGHM16}, which has been widely adopted for evaluating machine learning models under cross-device generalization settings~\citep{YLSWM26}. 
The dataset contains $640$~recordings collected from five individual E-Nose units (B1--B5), each equipped with an array of eight MOX sensors ($m=8$), including TGS2611, TGS2612, TGS2610, and TGS2602 sensors. 
Each unit was exposed to four gases ($n=4$) at ten concentration levels: $12.5$--$125.0$~ppm for ethanol and ethylene, and $25.0$--$250.0$~ppm for methane and carbon monoxide. 
For each session, responses from the eight MOX sensors were recorded for $600$~s at $100$~Hz. 

Following the protocol in~\citep{YLSWM26}, we use only the first $300$~s of each recording, as the second half corresponds to the purge stage of the sensing chamber. 
The selected window is then downsampled to $300$ time points for model input ($\tau=300$).
Although the five sensing units were designed to be identical, hardware variations and differences in data collection days introduced device-dependent drifts in the collected signals. 
While this study does not focus on drift adaptation, such variations offer a useful testbed for evaluating the generalizability of learned embeddings across different sensor conditions.

\subsection{Implementation Details \& Baselines}

We evaluate different models as the encoders~$\phi$, while using the same prediction heads for the classifier~$\mathcal{F}$ and regressor~$\mathcal{G}$. 
The classifier comprises a fully connected layer with $512$ nodes, followed by ReLU activation, and a four-node output layer for gas-type classification. 
The regressor also consists of a fully connected layer followed by a one-dimensional output layer with sigmoid activation.
Since the concentration ranges differ across gas types, we apply min--max normalization to the concentration values for each gas type so that the normalized targets range from $0$ to $1$. 
The predicted concentration values are then re-scaled back to 
their original ppm ranges for evaluation.

To evaluate the generalizability of the extracted representations across sensing units, we construct four cross-device test settings by holding out one sensing unit, B1--B4, at a time as the test set. 
The data from the remaining units are used for training and validation with a $9{:}1$ split. 
Following~\citep{YLSWM26}, all methane recordings from B5 are discarded due to reported data corruption, and B5 is therefore not used as a held-out test unit.
Moreover, all reported results are averaged over three random training--validation splits for each held-out test unit.

For comparative evaluation, Chronos-2~\citep{Chronos-2} and MOMENT~\citep{moment} are compared with the following baseline encoders:
\begin{itemize}
    \item MLP: A multilayer perceptron consisting of two connected layers of $1,024$ and $512$ hidden nodes with ReLU activation. Each sample is flattened into an input vector. 
    \item DBFE~\citep{YLSWM26}: State-of-the-art convolution-based model designed to capture temporal dynamics and cross-sensor correlations in ``gas'' sensor data. Since no official implementation is available, we implemented it based on the manuscript. Meta learning-specific losses introduced in~\citep{YLSWM26} are excluded to focus only on representational capability of the learned embeddings. 
    \item MLP+Chronos-2: Representations from MLP and Chronos-2 under fine-tuning are concatenated. 
\end{itemize}

MLP and DBFE are selected to represent domain-specific encoders trained directly on E-Nose data. 
For TSFM-based encoders, we use Chronos-2 Small~\citep{Chronos-2} and MOMENT Small~\citep{moment} to obtain $512$-dimensional embeddings ($d=512$).
Each TSFM is tested under frozen and fine-tuned settings. 
All models are trained for up to $1$K~epochs with a learning rate of $10^{-4}$ by minimizing both the cross-entropy loss and mean squared error (MSE).  
Early stopping is applied if the validation MSE does not improve for $50$~consecutive epochs. 
The checkpoint with the lowest validation MSE is used for final evaluation. 
Our code and data splits will be released upon acceptance.

\subsection{Results}

\subsubsection{Gas-type Classification}

\autoref{tab:cls_results} reports the accuracy and F1 scores for gas-type classification. 
Overall, performance varies substantially across held-out sensing units, indicating distributional differences among the units. 
For example, MLP achieves the best performance on B1 and B2, with accuracy scores above $.896$, but its accuracy drops to $.517$ and $.258$ on B3 and B4, respectively. 
Similarly, DBFE, a state-of-the-art model for E-Nose data, does not consistently generalize to unseen sensing units when used only as an encoder without the meta-learning strategy for which it was originally designed.

More importantly, frozen \mbox{Chronos-2} and MOMENT also fail to offer consistently strong performance across the test sets. 
Yet, \mbox{Chronos-2} shows substantial improvement after \emph{fine-tuning}, achieving sample-weighted F1 scores that are $107\%$ and $28\%$ higher than frozen \mbox{Chronos-2} and MLP, respectively. 
This implies that \mbox{Chronos-2} representations become useful for gas-sensing tasks only after task-specific learning.

Furthermore, \mbox{MLP+Chronos-2} outperforms fine-tuned \mbox{Chronos-2} on all test units except B3, suggesting that TSFM-based representations can provide \emph{complementary} information when fused with embeddings learned by task-specific models.

\subsubsection{Concentration Prediction}

\autoref{tab:reg_results} reports the root mean squared error~(RMSE) and mean absolute percentage error~(MAPE) for concentration prediction. Since the regressor conditions its prediction on the gas-type classification output, models that did not achieve promising classification performance in~\autoref{tab:cls_results} are excluded from this evaluation.

A pattern similar to the classification results is observed. 
\mbox{MLP+Chronos-2} achieves the best overall performance, whereas MLP performs the worst among the evaluated models. 
For example, fine-tuned \mbox{Chronos-2} reduces the sample-weighted RMSE by $13\%$ compared with MLP, and \mbox{MLP+Chronos-2} further improves the result by $19\%$.

These results suggest that fine-tuned \mbox{Chronos-2} provides useful embeddings for concentration prediction. Its strong classification performance also contributes to regression accuracy, since the predicted gas class determines both the input condition to the regressor and the inverse scaling from the sigmoid-normalized output to the original ppm range.


\section{Conclusion \& Future Work}

This paper examined TSFMs as feature encoders for gas-sensing time-series data from E-Nose systems. 
The results show that pretrained TSFM representations are limited effectiveness, while fine-tuning substantially improves performance. 
In addition, combining TSFM-based representations with domain-specific encoders further improves gas-type identification and concentration prediction. 
Future work will explore larger TSFMs and more diverse gas-sensing datasets

\section*{Acknowledgment}

The authors thank Subeen Choi for conducting preliminary E-Nose research and assisting with manuscript editing. 
This work was partially supported by the NSF (2502025) and by the Georgia Peanut Commission (KSU-1-26/26).

{\small
    \ifx\usenatbib\undefined%
	\bibliographystyle{IEEEtran}%
    \else%
    \bibliographystyle{IEEEtranN}%
    \fi
	\bibliography{references}

@article{YLSWM26,
  title={MDFE-Net: A Meta-Learning Driven Dual-Branch Feature Extraction Network for E-Nose Sensor Drift Adaptation},
  author={Yang, Qilong and Liu, Jinxia and Shi, Yan and Wang, Yanwei and Men, Hong},
  journal={ACS sensors},
  volume={11},
  number={5},
  pages={4057--4067},
  year={2026},
  publisher={ACS Publications}
}

@article{Chronos-2,
  title={Chronos-2: From univariate to universal forecasting},
  author={Ansari, Abdul Fatir and Shchur, Oleksandr and K{\"u}ken, Jaris and Auer, Andreas and Han, Boran and Mercado, Pedro and Rangapuram, Syama Sundar and Shen, Huibin and Stella, Lorenzo and Zhang, Xiyuan and others},
  journal={arXiv preprint arXiv:2510.15821},
  year={2025}
}

@article{moment,
  title={Moment: A family of open time-series foundation models},
  author={Goswami, Mononito and Szafer, Konrad and Choudhry, Arjun and Cai, Yifu and Li, Shuo and Dubrawski, Artur},
  journal={arXiv preprint arXiv:2402.03885},
  year={2024}
}

@article{FFGHM16,
  title={Calibration transfer and drift counteraction in chemical sensor arrays using Direct Standardization},
  author={Fonollosa, J and Fernandez, Luis and Guti{\'e}rrez-G{\'a}lvez, Agust{\'\i}n and Huerta, Ram{\'o}n and Marco, Santiago},
  journal={Sensors and Actuators B: Chemical},
  volume={236},
  pages={1044--1053},
  year={2016},
  publisher={Elsevier}
}

@article{GCM23,
  title={TimeGPT-1},
  author={Garza, Azul and Challu, Cristian and Mergenthaler-Canseco, Max},
  journal={arXiv preprint arXiv:2310.03589},
  year={2023}
}

@article{ZSZVH25,
  title={Benchmarking a Time-Series Foundation Model (TimeGPT) for Real-World Forecasting Applications},
  author={Zhang, Xiao and Sridharan, Srinath and Zahrin, Nur Hakim Bin and Venkataraman, Narayan and Hiong, Goh Siang},
  journal={Machine Learning with Applications},
  pages={100801},
  year={2025},
  publisher={Elsevier}
}

@article{LDXFSHSSYY26,
  title={Mira: Medical time series foundation model for real-world health data},
  author={Li, Hao and Deng, Bowen and Xu, Chang and Feng, Zhiyuan and Schlegel, Viktor and Huang, Yu-Hao and Sun, Yizheng and Sun, Jingyuan and Yang, Kailai and Yu, Yiyao and others},
  journal={Advances in Neural Information Processing Systems},
  volume={38},
  pages={98657--98685},
  year={2026}
}

@article{ZFJS25,
  title={From detection to forecasting: utilizing time-series foundation models to anticipate defects in metal additive manufacturing},
  author={Zhang, Jiayi and Farbiz, Farzam and Jafary-Zadeh, Mehdi and Sing, Swee Leong},
  journal={Journal of Manufacturing Processes},
  volume={150},
  pages={1040--1052},
  year={2025},
  publisher={Elsevier}
}

@article{RSSPW25,
  title={How Effective are Large Time Series Models in Hydrology? A Study on Water Level Forecasting in Everglades},
  author={Rangaraj, Rahuul and Shi, Jimeng and Shirali, Azam and Paudel, Rajendra and Wu, Yanzhao and Narasimhan, Giri},
  journal={arXiv preprint arXiv:2505.01415},
  year={2025}
}

@inproceedings{SZFGS14,
  title={MOX-NW electronic nose for detection of food microbial contamination},
  author={Sberveglieri, Giorgio and Zambotti, Giulia and Falasconi, Matteo and Gobbi, Emanuela and Sberveglieri, Veronica},
  booktitle={SENSORS, 2014 IEEE},
  pages={1376--1379},
  year={2014},
  organization={IEEE}
}

@article{RB14,
  title={Ultra low power MOX sensor reading for natural gas wireless monitoring},
  author={Rossi, Maurizio and Brunelli, Davide},
  journal={IEEE Sensors Journal},
  volume={14},
  number={10},
  pages={3433--3441},
  year={2014},
  publisher={IEEE}
}
}

\end{document}